\documentclass[10pt,twocolumn,letterpaper]{article}

\usepackage{graphicx}
\usepackage{amsmath}
\usepackage{amssymb}
\usepackage{booktabs}

\usepackage{cvpr}              % To produce the CAMERA-READY version
\usepackage[dvipsnames]{xcolor}

\definecolor{cvprblue}{rgb}{0.21,0.49,0.74}
\usepackage[pagebackref,breaklinks,colorlinks,allcolors=cvprblue]{hyperref}
\usepackage{overpic}
\usepackage{float}
\usepackage{bm}
\usepackage{multicol}
\usepackage{multirow}
\usepackage{pifont}
\usepackage{color}
\usepackage{multirow}
\usepackage{overpic}
\usepackage{arydshln}
\usepackage{makecell}
\usepackage{times}
\usepackage{epsfig}
\usepackage[T1]{fontenc}
\usepackage{hhline}
\usepackage{xcolor}
\usepackage{pifont}
\usepackage{enumitem}
\usepackage{colortbl}
\usepackage{verbatim}
\usepackage{bm}
\usepackage{fancyhdr}
\usepackage[accsupp]{axessibility}
\def\paperID{1261} % *** Enter the Paper ID here
\def\confName{CVPR}
\def\confYear{2025}

\definecolor{darkpink}{rgb}{0.91, 0.33, 0.5}
\definecolor{blue}{rgb}{0, 0, 255}
\definecolor{pink}{rgb}{255, 0, 255}
\definecolor{red}{rgb}{255, 0, 0}
\definecolor{green}{rgb}{0.0, 0.5, 0.0}
\definecolor{orange}{rgb}{1.0, 0.647, 0.0}
\definecolor{mygray}{gray}{.94}

\newcommand\blfootnote[1]{%
  \begingroup
  \renewcommand\thefootnote{}\footnote{#1}%
  \addtocounter{footnote}{-1}%
  \endgroup
}

\title{
        \textbf{GREAT}: Geometry-Intention Collaborative Inference for \par Open-Vocabulary 3D Object Affordance Grounding     
}

\author{Yawen Shao$^{1}$, Wei Zhai$^{1,\dagger}$, Yuhang Yang$^{1}$, Hongchen Luo$^{2}$, Yang Cao$^{1}$, Zheng-Jun Zha$^{1}$\\
{$^{1}$~MoE Key Laboratory of Brain-inspired Intelligent Perception and Cognition,}\\
{University of Science and Technology of China}\qquad
{$^{2}$~Northeastern University} \\
\small{\texttt{\{shaoyawen@mail., wzhai056@, yyuhang@mail.\}ustc.edu.cn}}\\
\small{\texttt{luohongchen@ise.neu.edu.cn}} \qquad
\small{\texttt{\{forrest@, zhazj@\}ustc.edu.cn}}
}

\begin{document}
\maketitle
\begin{abstract}
\blfootnote{$\dagger$Corresponding Author.}
Open-Vocabulary 3D object affordance grounding aims to anticipate ``action possibilities'' regions on 3D objects with arbitrary instructions, which is crucial for robots to generically perceive real scenarios and respond to operational changes. Existing methods focus on combining images or languages that depict interactions with 3D geometries to introduce external interaction priors. However, they are still vulnerable to a limited semantic space by failing to leverage implied invariant geometries and potential interaction intentions. Normally, humans address complex tasks through multi-step reasoning and respond to diverse situations by leveraging associative and analogical thinking. In light of this, we propose \textbf{GREAT} (\textbf{G}eomet\textbf{R}y-int\textbf{E}ntion coll\textbf{A}bora\textbf{T}ive inference) for Open-Vocabulary 3D Object Affordance Grounding, a novel framework that mines the object invariant geometry attributes and performs analogically reason in potential interaction scenarios to form affordance knowledge, fully combining the knowledge with both geometries and visual contents to ground 3D object affordance. Besides, we introduce the \textbf{P}oint \textbf{I}mage \textbf{A}ffordance \textbf{D}ataset \textbf{v2} (\textbf{PIADv2}), the largest 3D object affordance dataset at present to support the task. Extensive experiments demonstrate the effectiveness and superiority of GREAT. The code and dataset are available at \href{https://yawen-shao.github.io/GREAT/}{https://yawen-shao.github.io/GREAT/}.
\end{abstract}   
\section{Introduction}
\label{sec:intro}
\begin{figure*}[t]
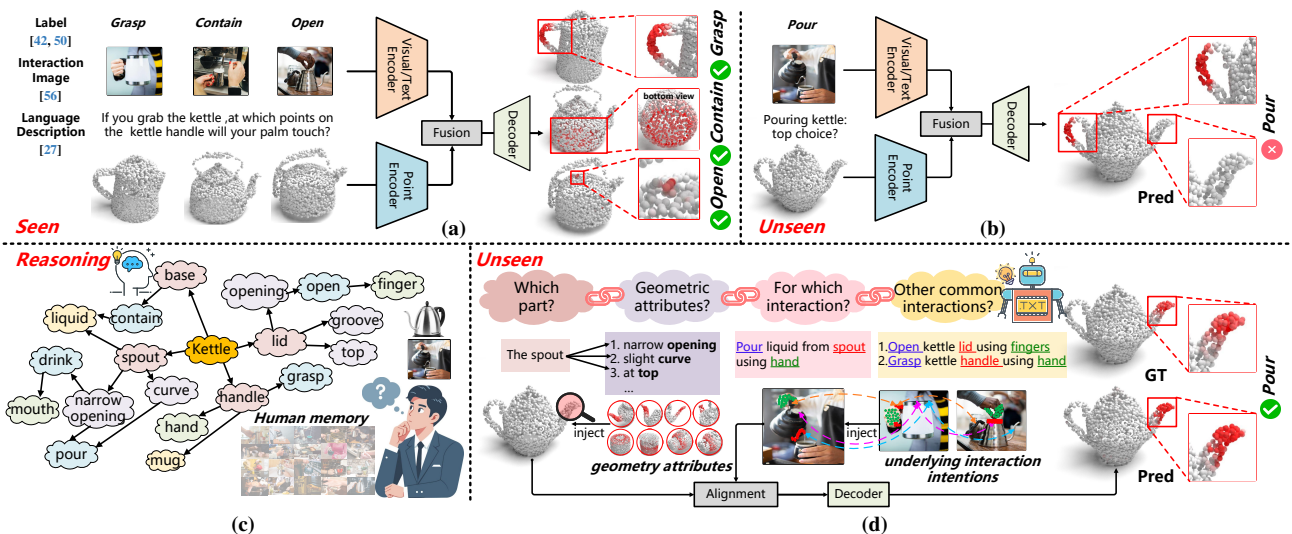

\centering
\small
\begin{overpic}[width=1\linewidth]{Figs/difference.pdf}
\put(34,21){{\textbf{\footnotesize (a)}}}
\put(75,21){{\textbf{\footnotesize (b)}}}
\put(18,-1.5){{\textbf{\footnotesize (c)}}}
\put(66,-1.5){{\textbf{\footnotesize (d)}}}
\put(2.65,35.5){{\textbf{\tiny \cite{Nguyen2023open,van2023open}}}}
\put(3.4,31.3){{\textbf{\tiny \cite{Yang_2023_ICCV}}}}
\put(3.4,27.2){{\textbf{\tiny \cite{LASO}}}}
\end{overpic}
\caption{\textbf{Difference and Motivation.} \textbf{(a)} Object affordance grounding on seen setting. \textbf{(b)} Open-Vocabulary Affordance Grounding (OVAG) with previous paradigms. \textbf{(c)} Observing interaction images, people engage in brainstorming through memory representations, drawing on prior interaction experiences to perform analogical reasoning and infer appropriate actions. \textbf{(d)} OVAG with our geometry-intention collaborative inference with chain-of-thought, step-by-step identifies the interaction part, extracts geometric attributes, reasons about corresponding interaction and brainstorms underlying interaction intentions, jointly grounding the 3D object affordance.}

\label{fig:difference}
\end{figure*}

Open-Vocabulary 3D object affordance grounding aims to locate ``action possibilities'' on objects  \cite{affordance, Hassanin2018VisualAA}, both for seen and unseen scenarios, identifying specific regions on objects that support certain interactions. This bridges visual perception and physical manipulation for embodied agents and possesses bountiful application scenarios, \eg robot manipulation \cite{huang2024rekep,kuang2024ramretrievalbasedaffordancetransfer,6225042}, scene understanding \cite{lcy,huang2023diffusion}, action anticipation \cite{zichen1,zichen2, ma2024madiff} and imitation learning \cite{li2024ag2manip, 10.1145/3054912}.

\par Recently, most existing methods \cite{deng20213d,6907679,partafford} establish explicit mappings between semantic affordance categories and geometric structures, restricted to predefined seen categories and fail to ground object affordance out of the training categories. Thus, some studies \cite{Yang_2023_ICCV,LASO,Nguyen2023open,van2023open, yang2024egochoir,zhaipami} explore grounding object affordance through additional instructions, encompassing combining images or languages that depict interactions with 3D geometries to introduce external interaction priors, and mitigate the generalization gap lead by affordance diversity. Despite their remarkable progress, they are still vulnerable to a limited semantic space by failing to leverage implied invariant geometries among objects with the same affordance, as well as potential correlations among distinct interactions of the same object. As shown in Fig. \ref{fig:difference} (a), current paradigms ground the 3D object affordance by aligning the object's geometric features and instruction modalities, working well in the seen partition. However, such a paradigm relies excessively on the data fed into the model, when testing unseen affordance like ``pour'' that does not appear in the training set, the model endeavors to categorize it as the seen category ``grasp'', shown in Fig. \ref{fig:difference} (b).

In cognitive science, studies \cite{Analogy,reason,zhai2022exploring} have shown that 
humans tackle complex tasks through multi-step reasoning and respond to diverse situations by employing associative and analogical thinking. As demonstrated in Fig. \ref{fig:difference} (c), when observing a water pouring scenario, humans employ multi-step reasoning to infer multiple potential interaction procedures and identify geometric properties of objects, indexing relevant knowledge from their brains. Analogous to this procedure, we leverage Multi-modal Large Language Models (MLLMs) \cite{internvl,li2022blipbootstrappinglanguageimagepretraining,mao2023gptdriverlearningdrivegpt,touvron2023llamaopenefficientfoundation} to simulate prior knowledge, encompassing visual, linguistic, and other modalities, lifting their superior reasoning and generalization abilities to visual tasks \cite{manipllm,peng2024globallocalcollaborativeinferencellm}. However, the dynamic and diverse nature of affordances makes it challenging to solely derive complex reasoning outcomes from MLLMs, to address this, a Chain-of-Thought (CoT) inference strategy that mirrors human reasoning processes is designed (Fig. \ref{fig:difference} (d)). The strategy step-by-step identifies the interaction primitives, extracts geometric attributes, and reasons about the corresponding interaction actions and intentions, which enables the model to analyze interaction images from multiple perspectives, thereby eliminating the hallucination and ambiguities raised by MLLMs when reasoning interactions.

\par To achieve this, we present the \textbf{GREAT}, a novel framework that excavates implied affordance knowledge from interaction images and effectively integrates it with both point cloud and image representations to jointly ground object 3D affordance. Specifically, we first devise a Multi-Head Affordance Chain-of-Thought (MHACoT) reasoning strategy to 
infer implied invariant geometries and underlying interaction intentions from fine-tuning MLLM, then GREAT employs attention mechanisms to model the correlation between these primitives to form the affordance knowledge dictionary. Following this, we design the Cross-Modal Adaptive Fusion Module (CMAFM) to integrate knowledge into the point cloud features and directly fuse it with image features, leveraging the combined representation to accurately ground 3D object affordance.

\par Moreover, to further unleash the model's capability, we extend the \textbf{P}oint \textbf{I}mage \textbf{A}ffordance \textbf{D}ataset (\textbf{PIAD}) \cite{Yang_2023_ICCV} to \textbf{PIADv2}, including 24 common affordances, 43 different object categories, over 15\textit{K} interaction images from diverse scenes and 38\textit{K} 3D objects with affordance annotations, \textbf{triple} interaction images and over \textbf{5 times} 3D instances compared to the PIAD (Tab. \ref{table:data}). 
%With its comprehensive scope, PIADv2 offers a robust and reliable testbed for 3D affordance grounding.

\begin{table}
\centering
\footnotesize
\renewcommand{\arraystretch}{1.}
\renewcommand{\tabcolsep}{6.pt}
\caption{\textbf{Comparison of related datasets.} Img.: interaction images. {$\sharp$ Img.}: number of images. {$\sharp$ 3D.}: number of 3D object instances. {$\sharp$ Obj.}: number of object categories. {$\sharp$ Aff.}: number of affordance categories.}
\label{table:data}

\begin{tabular}{c|ccccc}
\toprule
\textbf{Dataset} & \textbf{Img.} & \textbf{$\sharp$ Img.} & \textbf{$\sharp$ 3D.} & \textbf{$\sharp$ Obj.}  & \textbf{$\sharp$ Aff.}          \\ \midrule
3D AffordanceNet \cite{deng20213d} & \ding{55}  & $-$ & 22949 & 23 & 18  \\ 
PIAD \cite{Yang_2023_ICCV}     &  $\surd $ & 5162  & 7012 & 23 & 17    \\ 
\rowcolor{mygray} 
PIADv2 (Ours)     &  $\surd$ & 15213  & 38889 & 43 & 24   \\
\bottomrule
\end{tabular}
\vspace{-10pt}
\end{table}

\par The contributions are summarized as follows: 
\begin{itemize}[leftmargin=15pt,topsep=0pt,itemsep=2pt]
    \item[\textbf{1)}] We propose grounding 3D object affordance in an Open-Vocabulary fashion, which further reasons from interaction images, extrapolating from predefined sample space and generalize to unseen scenarios.
    \item[\textbf{2)}] We present GREAT, a novel framework that designs a MHACoT fine-tuning and inference strategy that excavates geometric attributes and underlying interaction intention to support the object affordance reasoning.
    \item[\textbf{3)}] We introduce the large-scale PIADv2 dataset, including 24 affordance and 43 object categories, 15\textit{K} interaction images from diverse scenes and over 38\textit{K} 3D objects with annotations. Extensive experiments on it demonstrate the effectiveness and superiority of GREAT.
\end{itemize}

\section{Related Work}
\label{sec:related work}

\textbf{Affordance Grounding.} 
Affordance grounding aims to locate the region of ``action possibilities'', which is a link between robot perception and manipulation. Some works ground the object affordance from the 2D data \eg images and videos \cite{Luo_2023_CVPR, Oneluo, zhai2023background, luo2021one,luotnnls,li:ooal:2024}, while some works leverage natural language understanding to ground affordance regions in 2D data \cite{Lu2022Phrase, li:ooal:2024, worldaff, qian2024affordancellm}. However, robot manipulation usually requires 3D information of objects, and the 2D affordance grounding obtained from the above works make it difficult to infer the interaction position of 3D objects. 
With the presentation of several 3D object datasets \cite{objaverse,Mo_2019_CVPR, sun2022benchmarking}, some works focus on the 3D object affordance grounding \cite{deng20213d,mo2021o2oafford, Mo_2019_CVPR}, which map semantic affordance to 3D object structure and fail to handle the open-vocabulary scenario.

\noindent\textbf{Open-Vocabulary 3D Affordance Grouding.} OVAG presents a substantial challenge, aiming to locate object affordance region in arbitrary scenario. Recently, some methods explore the possibility of OVAG. IAGNet \cite{Yang_2023_ICCV} utilizes the 2D interaction semantics to guide the grounding of 3D object affordance. LASO \cite{LASO} employs
textual-conditioned affordance queries to isolate afforded segments and injects text clues to point features. OpenAD \cite{Nguyen2023open} and OpenKD \cite{van2023open} propose a text-point correlation method for affordance synonym substitutions by utilizing the power of large language encoder clip. Although above methods have made remarkable progress, they are still vulnerable to a limited training semantic space due to the presence of critical learnable parts in the framework. GREAT mitigates this limitation 
by leveraging geometry-intention collaborative inference with CoT to ground object affordance.

\noindent\textbf{Chain-of-Thought Prompting with MLLMs.}
Chain-of-Thought (CoT) \cite{cot,zhang2022automaticchainthoughtprompting} and its variants \cite{yao2023treethoughtsdeliberateproblem,yao2024chainofthoughteffectivegraphofthoughtreasoning,lei2023boostinglogicalreasoninglarge} are proposed to enhance the reasoning capabilities of Multi-modal Large Language Models (MLLMs), which guide the model through multiple logical steps. With the rapid development of MLLMs \cite{internvl,wu2024visionllmv2,zhu2023minigpt}, vision-related \cite{yang2024kptllmunveilingpowerlarge,peng2024globallocalcollaborativeinferencellm,wu2024dettoolchainnewpromptingparadigm} methods have made significant progress in collaborating CoT and MLLMs to get the desired results. Some methods explore object affordance at task driven object detection \cite{Tang_2023_ICCV}, robot manipulation \cite{manipllm} and 2D-level object detection \cite{worldaff}, while these methods only describe the egocentric images and then reason through the text or image encoding results as inputs, that can only obtain limited and static knowledge. However, despite the considerable achievements of CoT enhanced MLLMs reasoning, it is challenging to reason about 3D object affordance from interaction images due to the complex and dynamic properties of affordance. To bridge this gap, we fine-tune the InternVL \cite{internvl} and directly input interaction images with prompts to reveal the geometric attributes and underlying interaction intentions.

\section{Method}

\subsection{Overview}
\label{Sec.3.1}
\begin{figure*}[t]
	\centering
        \scriptsize
	\begin{overpic}[width=1\linewidth]{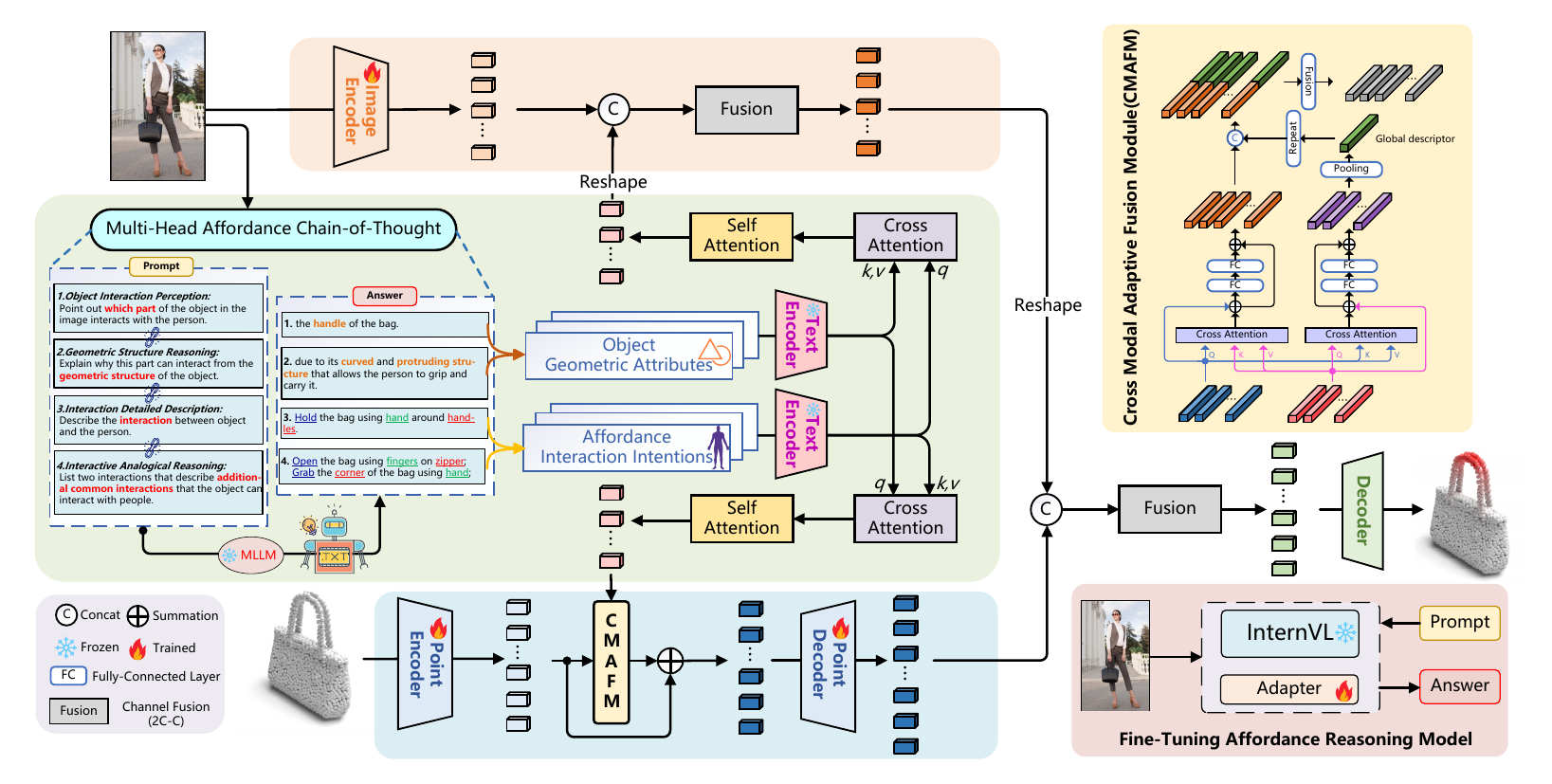}
        \put(19,2){\scriptsize$P$}
        \put(10,49){\scriptsize$I$}
        \put(33.5,44){\scriptsize$\mathbf{F}_i$}
        \put(35,8.5){\scriptsize$\mathbf{F}_p$}
        \put(57.5,44){\scriptsize$\mathbf{F}_{ti}$}
        \put(49.5,8.5){\scriptsize$\mathbf{P}_{o}$}
        \put(60.5,8.5){\scriptsize$\mathbf{F}_{tp}$}
        \put(54,29.2){\scriptsize$\mathbf{T}_{o}$}
        \put(54,22.8){\scriptsize$\mathbf{T}_{a}$}
        \put(36,16){\scriptsize$\bar{\mathbf{T}}_{o}$}
        \put(36,35){\scriptsize$\bar{\mathbf{T}}_{a}$}
        \put(84.1,18.2){\scriptsize$\mathbf{F}_{\alpha}$}
        \put(89.5,18.2){\scriptsize$\phi$}

        \put(90,46.5){\scriptsize$\mathbf{P}_{o}$}
        \put(80.5,23){\scriptsize$\mathbf{F}_p$}
        \put(87.5,23){\scriptsize$\bar{\mathbf{T}}_{o}$}
        
        \put(3,36){{\textbf{\footnotesize (a)}}}
        \put(92,23.5){{\textbf{\footnotesize (b)}}}
        \put(70,2.5){{\textbf{\footnotesize (c)}}}
	\end{overpic}
	\caption{\textbf{GREAT pipeline.} Initially, it extracts the respective features $\mathbf{F}_{i}, \mathbf{F}_{p}$ through modality-specific backbones, then results of MHACoT inference are encoded and aggregated to form object/affordance knowledge features $\bar{\mathbf{T}}_{o}, \bar{\mathbf{T}}_{a}$ (Sec. \ref{Sec.3.2}). Next, GREAT utilizes CMAFM to inject knowledge into $\mathbf{F}_{p}$ and $\mathbf{F}_{i}$ is directly fused to obtain fusion features $\mathbf{F}_{tp},\mathbf{F}_{ti}$ (Sec. \ref{Sec.3.3}). Eventually, $\mathbf{F}_{tp}$ and $\mathbf{F}_{ti}$ are sent to the decoder to ground 3D object affordance $\phi$ (Sec. \ref{Sec.3.4}).}
 \label{fig:method}
\end{figure*}
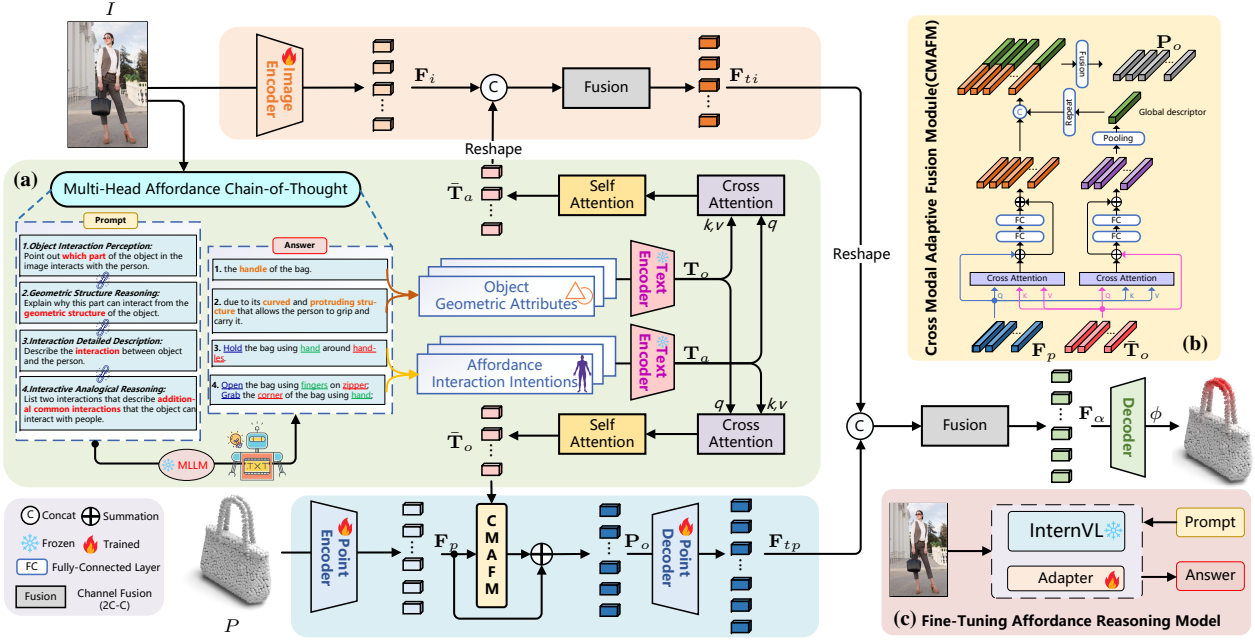

Given the inputs $\{P,I\}$, where $P \in \mathbb{R}^{N \times 4}$ denotes a point cloud of the object comprising the coordinate ${P_{c}} \in \mathbb{R}^{N \times 3}$ and the affordance annotation ${P_{label}} \in \mathbb{R}^{N \times 1}$, $I \in \mathbb{R}^{3 \times H \times W}$ denotes an image. $N$ is the number of points, $H, W$ are the height and width of an image. The goal is to optimize the model $f_{\theta}$ that outputs 3D object affordance $\phi$, expressed as $\phi = f_{\theta}(P_{c},I)$. As shown in Fig. \ref{fig:method}, initially, the inputs are sent to ResNet \cite{resnet} and PointNet++ \cite{qi2017pointnetplusplus}, obtain specific features $\mathbf{F}_{i} \in \mathbb{R}^{C \times H_{1} \times W_{1}}, \mathbf{F}_{p} \in \mathbb{R}^{C \times N_{p}}$, and reshape $\mathbf{F}_{i}$ to $\mathbb{R}^{C \times N_{i}}$ ($N_{i} = H_{1} \times W_{1}$). Then, GREAT captures object structure attributes and affordance interaction procedure by fine-tuning MLLM \cite{internvl} with the multi-head affordance
chain-of-thought to reason about interaction images. Next, the features encoded by Roberta \cite{liu2019robertarobustlyoptimizedbert} are fused through cross-attention mechanism, calculating object geometric feature $\bar{\mathbf{T}}_{o}$ and affordance intention feature $\bar{\mathbf{T}}_{a}$ (Sec. \ref{Sec.3.2}). Afterwards, with $\bar{\mathbf{T}}_{o}$, $\bar{\mathbf{T}}_{a}$ as knowledge dictionaries, GREAT leverages Cross-Modal Adaptive Fusion Module to inject knowledge clues into point features and directly fuse knowledge within image features to obtain fusion features $\mathbf{F}_{tp},\mathbf{F}_{ti}$ (Sec. \ref{Sec.3.3}). Eventually, $\mathbf{F}_{tp}$ and $\mathbf{F}_{ti}$ are sent to the decoder to ground 3D object affordance $\phi$, the whole process is optimized by a combined loss (Sec. \ref{Sec.3.4}). 

\subsection{Multi-Head Affordance Chain-of-Thought}
\label{Sec.3.2}
\textbf{Fine-Tuning MLLM.} We adopt InternVL \cite{internvl} and the injected learnable adapters \cite{hu2022lora} to fine-tune the MLLM, due to the MLLM primarily focuses on object recognition and description without sufficient understanding of what objects are actually used for and how they interact with humans. As shown in Fig. \ref{fig:method} (c), given an interaction image $I \in \mathbb{R}^{3 \times H \times W}$ and text prompts $T$, InternVL is desired to perform multi-modal understanding and give correct answers. During the training process, we only fine-tune the injected adapters for 10 epochs with a learning rate of 4e-5 and a LoRA rank of 16, while freezing the main parameters, to preserve the power of InternVL and further empower the model with capabilities in affordance understanding. 

\noindent\textbf{Object-Head Reasoning for Geometry.} It consists of \textit{Object Interaction Perception} and \textit{Geometric Structure Reasoning}. First, the model needs to focus on understanding the interactive components of an object within the image, refining its perception of the object's key parts rather than the entire object. As illustrated in Fig. \ref{fig:method} (a), we design the first prompt as ``\textit{Point out which part of the {object} in the image interacts with the person.}'' Next, since similar regions of different objects can often perform the same interaction based on shared geometric attributes, the model needs to reason about these features. This allows it to move beyond the constraints of object categories and focus more on the relationship between structure and affordance. Thus, we design the second prompt as ``\textit{Explain why this part can interact from the geometric structure of the {object}.}'' 

\noindent\textbf{Affordance-Head Reasoning as Brainstorming.} It consists of \textit{Interaction Detailed Description} and \textit{Interactive Analogical Reasoning }. First, the model needs to identify the entire interaction process between the object and the person in the image, including the interaction parts on both the object and the person, as well as the type of interaction. This allows the model to generate a fine-grained feature representation and capture the physical structure constraints of the interaction between the person and the object. As shown in Fig. \ref{fig:method} (a), we design the third prompt as ``\textit{Describe the interaction between {object} and the person.}'' Next, in the human mind, observing an object is typically followed by associating it with various potential ways of interaction. Inspired by this, the MLLM's world knowledge repository is leveraged to explore other possible interaction intentions of an object, reducing reliance on specific affordance instances and enhancing analogical reasoning. We design the fourth prompt as ``\textit{List two interactions that describe additional common interactions that the {object} can interact with people.}'' Due to the complexity of interaction images, {object} in the prompts is filled with the object category. Only the key part of the prompts is provided here, for the full prompts, please refer to the appendix.

\noindent\textbf{Knowledge Encoding and Integration.}
For each interaction image, we concatenate the textual descriptions of the interaction components and the object's geometric attributes inferred from the object-head into a textual sequence. Additionally, we combine the actual interaction and two potential interactions inferred from the affordance-head. The text encoder Roberta \cite{liu2019robertarobustlyoptimizedbert} is used to obtain object geometric knowledge feature $\mathbf{T}_{o} \in \mathbb{R}^{N_{o} \times C}$ and affordance intention knowledge feature $\mathbf{T}_{a} \in \mathbb{R}^{N_{a} \times C}$, where $N_{o}, N_{a}$ denote as the number of interaction objects and the number of interaction ways. By applying the cross-attention layer $f_m$ to correlate information from the two knowledge repositories and the self-attention layer $f_{\delta}$ to enrich the contextual information, the process aligns the object's geometric structure attributes with potential interaction intentions, as formulated below:
\begin{equation}
\small
\label{Eq:1}
    \bar{\mathbf{T}}_{o} = f_{\delta}(f_m(\mathbf{T}_{o}, \mathbf{T}_{a})), \bar{\mathbf{T}}_{a} = f_{\delta}(f_m(\mathbf{T}_{a}, \mathbf{T}_{o})),
\end{equation}
where $\bar{\mathbf{T}}_{o} \in \mathbb{R}^{N_{o} \times C}$, $\bar{\mathbf{T}}_{a} \in \mathbb{R}^{N_{a} \times C}$.

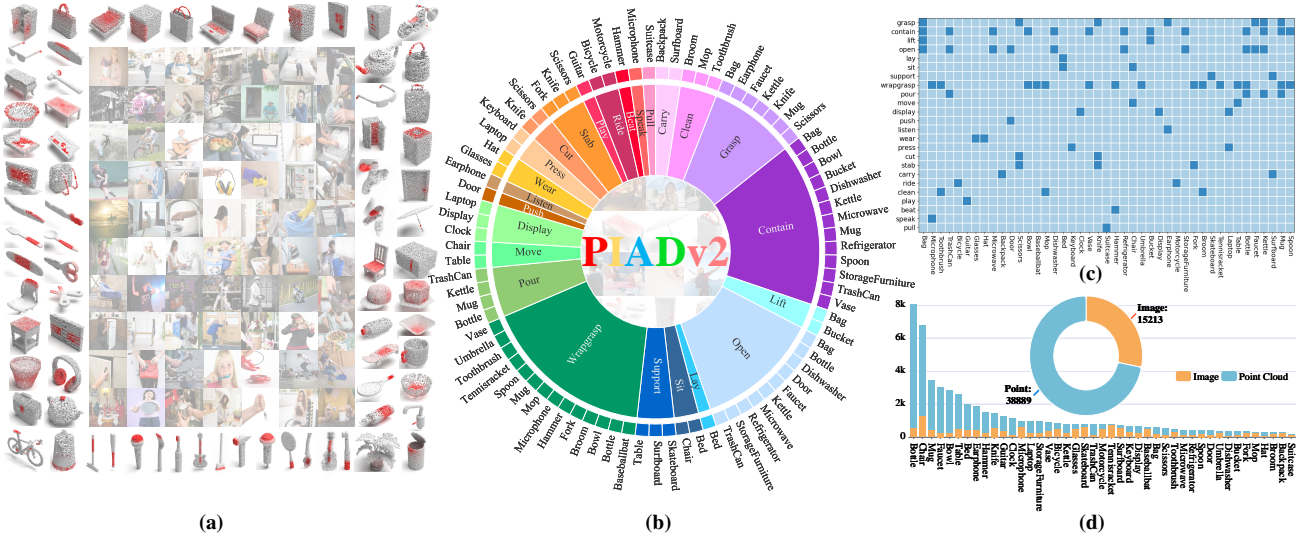
\begin{figure*}[t]
        \centering
	\small
        \begin{overpic}[width=1\linewidth]{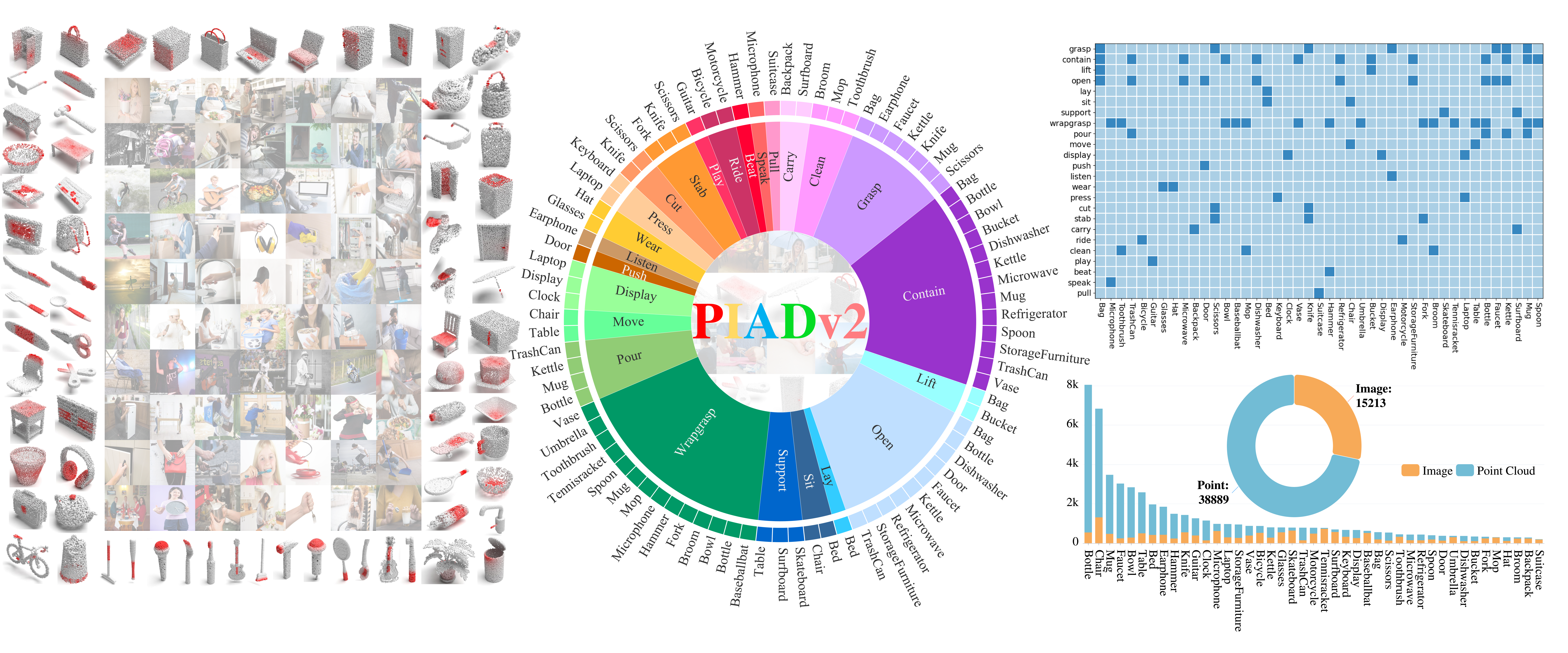}
        \put(15,0){{\textbf{\footnotesize (a)}}}
        \put(49,0){{\textbf{\footnotesize (b)}}}
        \put(82,19){{\textbf{\footnotesize (c)}}}
        \put(82,0){{\textbf{\footnotesize (d)}}}
        \end{overpic}
        \caption{\textbf{PIADv2 Dataset.} \textbf{(a)} Extensive data examples from PIADv2, the red region in point clouds is the affordance annotation. \textbf{(b)} Category distribution in PIADv2. \textbf{(c)} Confusion matrix between affordance and object categories, where the horizontal axis represents object category and the vertical axis represents affordance category. \textbf{(d)} Ratio of images and point clouds in each object category.}
 \label{Fig:dataset}
\end{figure*}

\subsection{Cross-Modal Adaptive Fusion Module}
\label{Sec.3.3}
To better facilitate the cross-modal fusion of the geometric attributes of interaction regions and point cloud features, we propose Cross-Modal Adaptive Fusion Module (CMAFM) that integrates the geometric attributes into the deepest encoder layer of PointNet++ \cite{qi2017pointnetplusplus} to refine the point feature map, enabling effective cross-modal feature alignment and fusion, as shown in Fig. \ref{fig:method} (b).

Specifically, CMAFM re-represents $\mathbf{F}_{p} \in \mathbb{R}^{C \times N_{p}}$ and $\bar{\mathbf{T}}_{o} \in \mathbb{R}^{N_{o} \times C}$ in the same feature space to further align the local features. $\mathbf{F}_{p}$ is projected to form the query $\mathbf{Q} = \mathbf{F}_{p}\mathbf{W}_{1}$, $\bar{\mathbf{T}}_{o}$ is projected to form the key $\mathbf{K} = \bar{\mathbf{T}}_{o}\mathbf{W}_{2}$ and value $\mathbf{V} = \bar{\mathbf{T}}_{o}\mathbf{W}_{3}$, where $\mathbf{W}_{1\sim 3}$ are projection weights. The cross-attention mechanism integrates these features to extract the interaction context information, formulated as:
\begin{equation}
\small
\mathbf{F}^{'}_{p} = (softmax(\mathbf{Q}^{T} \cdot \mathbf{K}/ \sqrt{d}) \cdot \mathbf{V}^{T})^{T},
\end{equation}
where $\mathbf{Q} \in \mathbb{R}^{d \times N_{p}}, \mathbf{K}, \mathbf{V} \in \mathbb{R}^{d \times N_{o}}, \mathbf{F}^{'}_{p} \in \mathbb{R}^{C \times N_{p}} $, $d$ is the dimension of projection. Similarly $\bar{\mathbf{T}}^{'}_{o} \in \mathbb{R}^{C \times N_{o}}$ can be re-represented. Then, CMAFM injects object geometric structure attributes into each point in the point cloud features to obtain the fused point features ${\mathbf{P}}_{o}$, formulated as:
\begin{equation}
\small
\mathbf{P}_{o} = f[ \mathbf{F}^{'}_{p} + f_{\varphi}(\mathbf{F}^{'}_{p}), \Theta ( \bar{\mathbf{T}}^{'}_{o} + f_{\varphi}(\bar{\mathbf{T}}^{'}_{o}))],
\end{equation}
where $f_{\varphi}$ denotes two Fully-Connected (FC) layers, $\Theta$ denotes pooling and expand it to $\mathbb{R}^{C \times N_{p}}$, $[\cdot]$ denotes the concatenation, $f$ indicates convolution layers with $1 \times 1$ kernel. Finally, $\mathbf{P}_{o}$ is upsampled to $\mathbb{R}^{C \times N}$ by Feature Propagation layers (FP) \cite{qi2017pointnetplusplus}, formulated as: $\mathbf{F}_{tp} = \mathbf{FP}(\mathbf{P}_{o})$.

To understand the multiple affordances of objects and provide rich contextual information for analogical reasoning, underlying interaction intention textual features are fused with the image features:
\begin{equation}
\small
\mathbf{F}_{ti} = f[\Gamma(\bar{\mathbf{T}}_{a}), \mathbf{F}_{i}], \mathbf{F}_{ti} \in \mathbb{R}^{C \times N_{i}},
\end{equation}
where $\Gamma$ denotes reshape $\bar{\mathbf{T}}_{a}$ to $\mathbb{R}^{C \times N_{i}}$.

\subsection{Decoder and Loss Functions}
\label{Sec.3.4}
Image features with interaction intention and point features with geometric structure are fed into the decoder, which jointly reveals the 3D affordance region, formulated as:
\begin{equation}
\small
\mathbf{F}_{\alpha} = f[\Gamma({\mathbf{F}}_{ti}), \mathbf{F}_{tp}], \phi = \sigma(f_{\phi}(\mathbf{F}_{\alpha})),
\end{equation}
where $\Gamma$ denotes reshape ${\mathbf{F}}_{ti}$ to $\mathbb{R}^{C \times N}$, $f_{\phi}$ denotes an output head, $\sigma$ denotes the sigmoid function, $\mathbf{F}_{\alpha} \in \mathbb{R}^{C \times N}$ is affordance feature representation and $\phi \in \mathbb{R}^{N \times 1}$ represents the 3D object affordance.

Unconstrained by the affordance category labels, we focus on the differences between 3D object affordance $\phi$ and affordance ground truth annotation $P_{label}$, enabling the model to directly link 3D object affordances with interaction images through reasoning. Therefore, the total loss consists of a focal loss \cite{focal_loss} and a dice loss \cite{dice_loss}, which supervises point-wise heatmaps, formulated as:
\begin{equation}
\small
\mathbf{\mathcal{L}}_{total} = \mathbf{\mathcal{L}}_{focal} + \mathbf{\mathcal{L}}_{dice}.
\end{equation}

\section{Dataset}
\begin{table*}[t]
\footnotesize
\centering
  \renewcommand{\arraystretch}{1.}
  \renewcommand{\tabcolsep}{0.5pt}
  \caption{\textbf{Comparison on the PIADv2.} Evaluation metrics of comparison methods on the benchmark, the best results are in \textbf{bold}. \textbf{Seen}, \textbf{Unseen Object} and \textbf{Unseen Affordance} are three partitions of the dataset. AUC and aIOU are shown in percentage. $\textcolor{darkpink}{\diamond}$ denotes the relative improvement of our method over other methods.}
\label{table:main_results}
\begin{tabular}{c|cccc|cccc|cccc}
\toprule
\multicolumn{1}{c|}{}   &\multicolumn{4}{c|}{\textbf{Seen}}                                                                     & \multicolumn{4}{c|}{\textbf{Unseen Object}}                                            & \multicolumn{4}{c}{\textbf{Unseen Affordance}}                  \\ \midrule
\multicolumn{1}{c|}{\textbf{Methods}} & \textbf{AUC $\uparrow$} & \textbf{aIOU $\uparrow$} & \textbf{SIM $\uparrow$} & \textbf{MAE$\downarrow$} & \textbf{AUC $\uparrow$} & \textbf{aIOU $\uparrow$} & \textbf{SIM $\uparrow$} & \textbf{MAE$\downarrow$} & \textbf{AUC $\uparrow$} & \textbf{aIOU $\uparrow$} & \textbf{SIM $\uparrow$} & \textbf{MAE$\downarrow$}\\ \midrule
\multicolumn{1}{c|}{Baseline}            
& $87.04$              & $34.18$           & $0.594$       & $0.079$
& $72.74\textcolor{darkpink}{\scriptstyle~\diamond9.4\%}$            & $16.34\textcolor{darkpink}{\scriptstyle~\diamond23.4\%}$           &
$0.336\textcolor{darkpink}{\scriptstyle~\diamond19.6\%}$       & 
$0.156\textcolor{darkpink}{\scriptstyle~\diamond30.1\%}$       
& $58.09\textcolor{darkpink}{\scriptstyle~\diamond20.2\%}$              & $7.88\textcolor{darkpink}{\scriptstyle~\diamond52.9\%}$            & 
$0.208\textcolor{darkpink}{\scriptstyle~\diamond39.4\%}$       & 
$0.160\textcolor{darkpink}{\scriptstyle~\diamond20.6\%}$ \\
\multicolumn{1}{c|}{FRCNN \cite{frcnn}}            
& $87.05$              & $33.55$           & $0.600$       & $0.082$
& $72.20\textcolor{darkpink}{\scriptstyle~\diamond10.2\%}$             & $18.08\textcolor{darkpink}{\scriptstyle~\diamond11.5\%}$           & 
$0.362\textcolor{darkpink}{\scriptstyle~\diamond11.0\%}$       &
$0.152\textcolor{darkpink}{\scriptstyle~\diamond28.3\%}$       
& $59.08\textcolor{darkpink}{\scriptstyle~\diamond18.2\%}$              & $7.96\textcolor{darkpink}{\scriptstyle~\diamond51.4\%}$            & 
$0.210\textcolor{darkpink}{\scriptstyle~\diamond38.1\%}$       & 
$0.156\textcolor{darkpink}{\scriptstyle~\diamond18.6\%}$ \\  
\multicolumn{1}{c|}{XMF \cite{xmf}}            
& $87.39$              & $33.91$           & $0.604$       & $0.078$
& $74.61\textcolor{darkpink}{\scriptstyle~\diamond6.6\%}$              & 
$17.40\textcolor{darkpink}{\scriptstyle~\diamond15.9\%}$           & 
$0.361\textcolor{darkpink}{\scriptstyle~\diamond11.4\%}$       & 
$0.126\textcolor{darkpink}{\scriptstyle~\diamond13.5\%}$       & 
$60.99\textcolor{darkpink}{\scriptstyle~\diamond14.5\%}$              & 
$8.11\textcolor{darkpink}{\scriptstyle~\diamond48.6\%}$            & 
$0.225\textcolor{darkpink}{\scriptstyle~\diamond28.9\%}$       & 
$0.152\textcolor{darkpink}{\scriptstyle~\diamond16.4\%}$ \\ 
\multicolumn{1}{c|}{IAG \cite{Yang_2023_ICCV}}            
& $89.03$              & $34.29$           & $0.623$       & $0.076$
& $73.03\textcolor{darkpink}{\scriptstyle~\diamond9.0\%}$              & $16.78\textcolor{darkpink}{\scriptstyle~\diamond20.1\%}$           & 
$0.351\textcolor{darkpink}{\scriptstyle~\diamond14.5\%}$       & 
$0.123\textcolor{darkpink}{\scriptstyle~\diamond11.4\%}$      & 
$62.29\textcolor{darkpink}{\scriptstyle~\diamond12.1\%}$              & $8.99\textcolor{darkpink}{\scriptstyle~\diamond34.0\%}$            & 
$0.251\textcolor{darkpink}{\scriptstyle~\diamond15.5\%}$       & 
$0.141\textcolor{darkpink}{\scriptstyle~\diamond9.9\%}$ \\ 
\multicolumn{1}{c|}{LASO \cite{LASO}}            
& $90.34$              & $34.88$           & $0.627$       & $0.077$
& $73.32\textcolor{darkpink}{\scriptstyle~\diamond8.5\%}$              & $16.05\textcolor{darkpink}{\scriptstyle~\diamond25.6\%}$           & 
$0.354\textcolor{darkpink}{\scriptstyle~\diamond13.6\%}$       & 
$0.123\textcolor{darkpink}{\scriptstyle~\diamond11.4\%}$       
& $64.07\textcolor{darkpink}{\scriptstyle~\diamond9.0\%}$              & $8.37\textcolor{darkpink}{\scriptstyle~\diamond44.0\%}$            & 
$0.228\textcolor{darkpink}{\scriptstyle~\diamond27.2\%}$       & 
$0.140\textcolor{darkpink}{\scriptstyle~\diamond9.3\%}$ \\

\rowcolor{mygray} 
\multicolumn{1}{c|}{\textbf{Ours} }            
& $\textbf{91.99}$   & $\textbf{38.03}$  & $\textbf{0.676}$  & $\textbf{0.067}$  
& $\textbf{79.57}$   & $\textbf{20.16}$  & $\textbf{0.402}$  & $\textbf{0.109}$
& $\textbf{69.81}$   & $\textbf{12.05}$  & $\textbf{0.290}$  & $\textbf{0.127}$
\\ 
\bottomrule
\end{tabular}
\end{table*}

\noindent\textbf{Collection.} We construct the \textbf{P}oint \textbf{I}mage \textbf{A}ffordance \textbf{D}ataset \textbf{v2} (\textbf{PIADv2}), which comprises paired 2D interaction images and 3D object point clouds. Points are mainly collected from 3DIR \cite{yang2024lemon}, 3D-AffordanceNet \cite{deng20213d}, objaverse \cite{objaverse}, etc. Images are  mainly collected from AGD20k \cite{Learningluo}, OpenImage \cite{openimages} and websites with free licenses. In total, PIADv2 contains 15213 images and 38889 point clouds, covering 43 object and 24 affordance categories, which is the largest scale 3D object affordance dataset so far. Some data samples are shown in Fig. \ref{Fig:dataset} (a). The affordance and the object categories are shown in Fig. \ref{Fig:dataset} (b). Both figures show that the dataset covers numerous affordances, supporting various interaction scenarios and diverse object categories.

\noindent\textbf{Annotation.} We annotate the affordance of each point cloud instance by affordance category. For instance, an annotation is a matrix of (2048, 4), 2048 is the number of points, and $4$ indicates 3D coordinates with the corresponding affordance heatmap, each affordance category possesses such annotation of an instance. For images, we classify images according to the affordance category. 

% when an image is ambiguous, we demand annotators to categorize the image according to the main affordance shown in the image, \eg pouring water with kettle contains both pour and grasp, we define it as pour, and only holding a kettle without pouring water is defined as grasp. Some 3D annatation examples are shown in Fig. \ref{Fig:dataset} (a). 

\noindent\textbf{Statistical Analysis.} 
In our training setting, images and point clouds do not require a fixed one-by-one pairing, as they are sampled from different instances to ensure the generalization to distinct instances. Fig. \ref{Fig:dataset} (c) shows the confusion matrix of affordance and object categories, revealing a multi-to-multi relationship, which poses a significant challenge to the accuracy and generalization of the 3D object affordance grounding. Fig. \ref{Fig:dataset} (d) shows the ratio of images and point clouds for each object category, offering insights into the balance between interaction images and 3D object point clouds, further highlighting its comprehensiveness.

\noindent\textbf{Data Partitions.}
Our dataset provides three partitions, two of which follow the PIAD \cite{Yang_2023_ICCV}. \textbf{Seen}: the training and test sets share the same objects and affordances. \textbf{Unseen Object}: affordances are consistent between the training and test sets, but some objects in the test set do not appear in the training set. \textbf{Unseen Affordance}: affordances in the test set are not present in the training set, and so does certain objects. Please refer to the appendix for more details. 
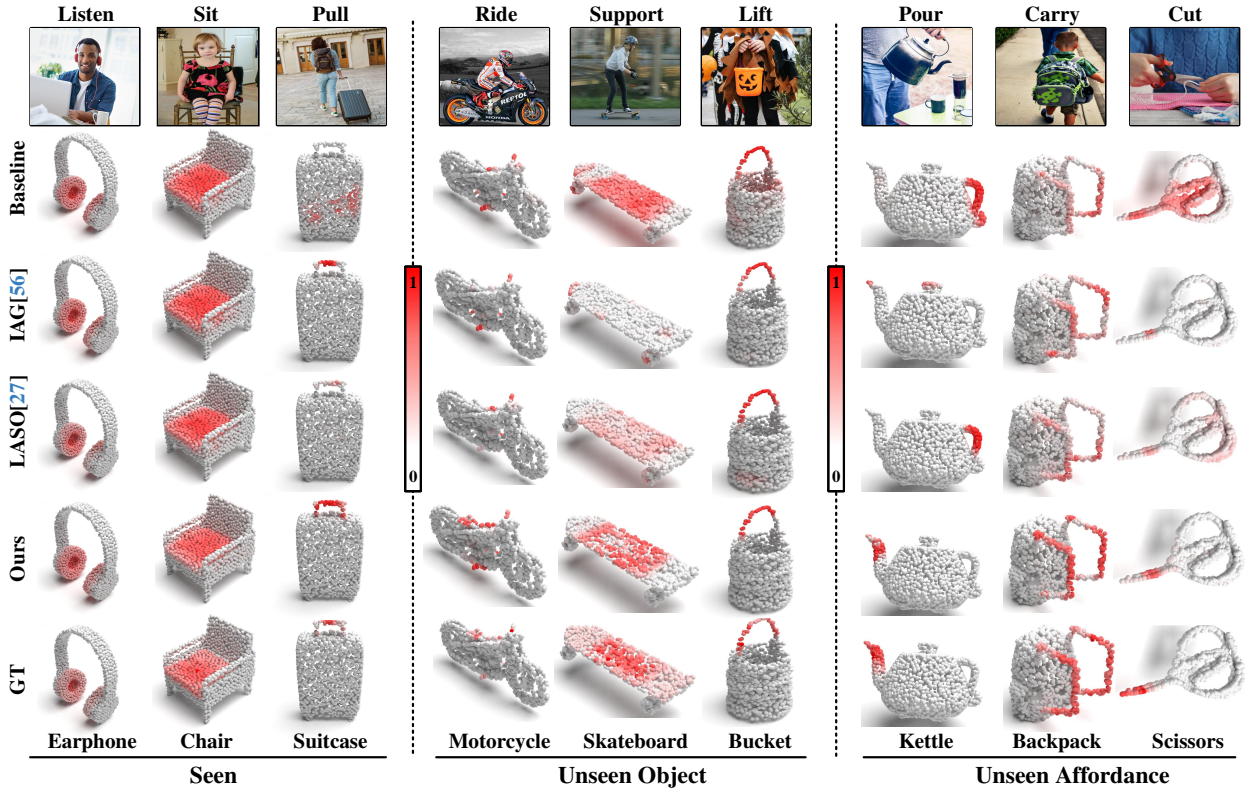
\begin{figure*}[t]
	\centering
	\small
        \begin{overpic}[width=0.96\linewidth]{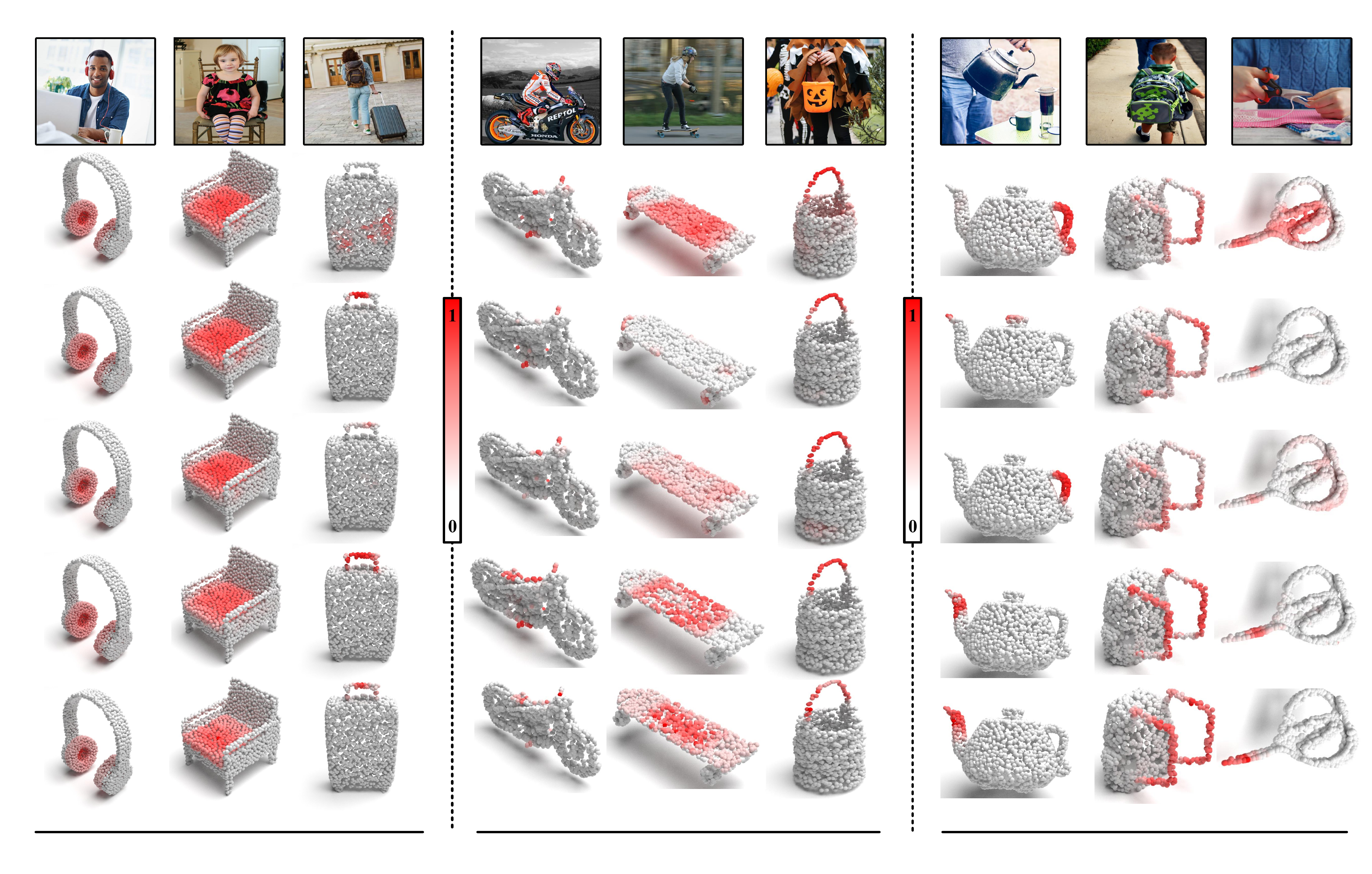}
        \put(1,8){\rotatebox{90}{\textbf{\footnotesize GT}}}
        \put(1,17){\rotatebox{90}{\textbf{\footnotesize Ours}}}
        \put(1,26){\rotatebox{90}{\textbf{\footnotesize LASO\cite{LASO}}}}
        \put(1,36){\rotatebox{90}{\textbf{\footnotesize IAG\cite{Yang_2023_ICCV}}}}
        \put(1,46){\rotatebox{90}{\textbf{\footnotesize Baseline}}} 

        \put(4.8,61.5){\textbf{\footnotesize Listen}}
        \put(15.6,61.5){\textbf{\footnotesize Sit}}
        \put(25,61.5){\textbf{\footnotesize Pull}}
        \put(37.9,61.5){\textbf{\footnotesize Ride}}
        \put(47,61.5){\textbf{\footnotesize Support}}
        \put(58.8,61.5){\textbf{\footnotesize Lift}}
        \put(71.5,61.5){\textbf{\footnotesize Pour}}
        \put(81.5,61.5){\textbf{\footnotesize Carry}}
        \put(92.85,61.5){\textbf{\footnotesize Cut}}

        \put(4,3.8){\textbf{\footnotesize Earphone}}
        \put(14.5,3.8){\textbf{\footnotesize Chair}}
        \put(23.5,3.8){\textbf{\footnotesize Suitcase}}
        \put(35.8,3.8){\textbf{\footnotesize Motorcycle}}
        \put(46.5,3.8){\textbf{\footnotesize Skateboard}}
        \put(58,3.8){\textbf{\footnotesize Bucket}}
        \put(71.5,3.8){\textbf{\footnotesize Kettle}}
        \put(80.5,3.8){\textbf{\footnotesize Backpack}}
        \put(91.6,3.8){\textbf{\footnotesize Scissors}}

        \put(14.8,1){\textbf{ Seen}}
        \put(44,1){\textbf{ Unseen Object}}
        \put(77.1,1){\textbf{ Unseen Affordance}}
        
	\end{overpic}
	\caption{\textbf{Visualization Results.} The first row is the interaction image and the last row is the ground truth of 3D object affordance in point cloud. The left-middle-right partitions correspond to the visual comparison results for different 3D object affordance in the Seen, Unseen Object, and Unseen Affordance partitions, respectively. The depth of \textcolor[rgb]{0.83,0.24,0.21}{red} represents the affordance probability.}
 \label{Fig:mainresults}
\end{figure*}
\section{Experiment}
\subsection{Benchmark Setting}
\textbf{Evaluation Metrics.}
For a thorough assessment, we use previous evaluation metrics from advanced works \cite{Yang_2023_ICCV, LASO} to benchmark the model on our PIADv2 dataset, which include \textbf{A}rea \textbf{U}nder \textbf{C}urve \cite{Lobo2008AUCAM} , \textbf{a}verage \textbf{I}ntersection \textbf{O}ver \textbf{U}nion \cite{iou}, \textbf{SIM}ilarity \cite{sim}, \textbf{M}ean \textbf{A}bsolute \textbf{E}rror \cite{mae}. 

\par\noindent\textbf{Compare Baselines.}
For a comprehensive comparison, we select two leading works (IAG \cite{Yang_2023_ICCV} and LASO \cite{LASO}) on 3D affordance grounding and two leading image-point cloud cross-modal works (FRCNN \cite{frcnn} and XMF \cite{xmf}) mentioned in the IAG. Following IAG, the baseline directly concatenates the features from modality-specific extractors.

\noindent\textbf{Implementation Details.}
We take PointNet++ \cite{qi2017pointnetplusplus} and ResNet18 \cite{resnet} as the default 3D and 2D backbones, respectively. To ensure a fair comparison, the same feature extractor and settings are used to reproduce the baselines. We train the GREAT for 65 epochs with a batch size of 16, utilizing the Adam optimizer with a learning rate of 1e-4. 
%More details about the benchmark setting can be found in the appendix. 

\subsection{Comparison Results}
The comparison results of evaluation metrics are presented in Tab. \ref{table:main_results}, demonstrating \textbf{GREAT} significantly outperforms the compared baselines across all metrics in three partitions and achieves the state-of-the-art performance. Furthermore, the results are visualized in Fig. \ref{Fig:mainresults}.

\noindent\textbf{Seen vs. Unseen.}
Quantitatively analyzing the results of the Tab. \ref{table:main_results}, all baselines demonstrate a stepwise metrics decrease in all partitions. This trend emphasizes the difficulty of generalizing unseen objects and affordances. Compared to other baselines, the superior performance demonstrated by GREAT in the open-vocabulary proves the necessity and rationality of our task setting for open-vocabulary affordance grounding.
Qualitative analysis of the visualization results in Fig. \ref{Fig:mainresults} shows little difference in the Seen setting, but in the Unseen setting, GREAT significantly outperforms the other methods. For example, in the case of kettle, only IAG detects the small affordance region for pouring, while methods that directly link object structure with textual descriptions fail to predict and can only identify the trained affordance: grasp. In contrast, our method uncovers interaction details in 2D interaction images by leveraging the MLLM embedded with world knowledge for MHACoT reasoning, leading to more precise predictions.

\subsection{Ablation Study}
We conduct a thorough ablation study to validate the effectiveness of the model design, as shown in Tab. \ref{table:ablation}. First, we ablate the object head and affordance head in MHACoT separately to demonstrate the importance of acquiring geometric attributes of objects and underlying interaction intentions. Additionally, removing the CMAFM module impairs the alignment between geometric attributes and the point cloud, while omitting MLLM fine-tuning reduces the accuracy of reasoning, resulting in decreased performance.

\begin{figure}
    \centering
    \small
    \begin{overpic}[width=1.\linewidth]{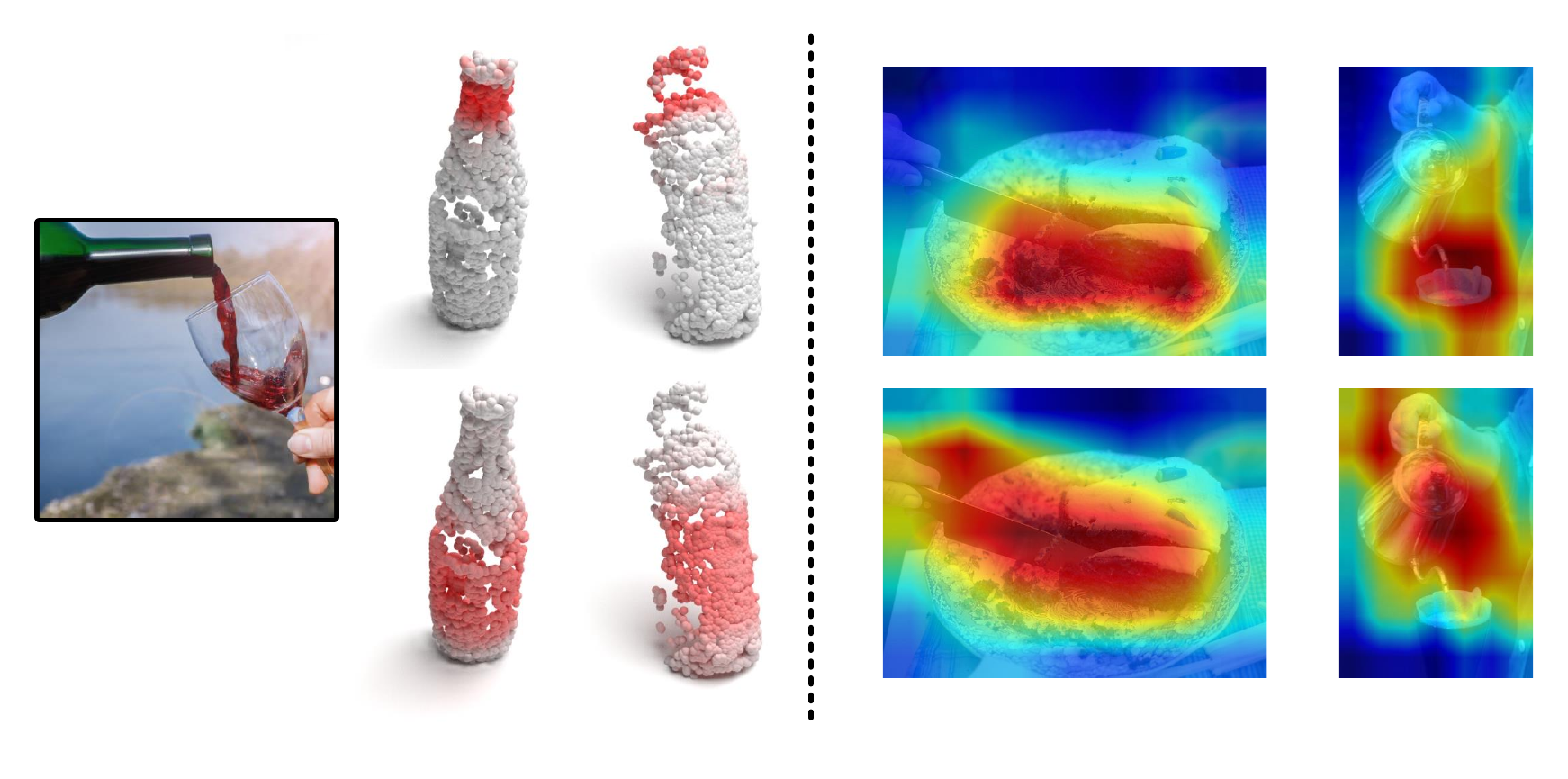}
    \put(29,49.5){\textbf{$\bm{w}$ AffCoT}}
    \put(27,0.5){\textbf{$\bm{w/o}$ AffCoT}}
    \put(72,49.5){\textbf{$\bm{w}$ ObjCoT}}
    \put(70,0.5){\textbf{$\bm{w/o}$ ObjCoT}}

    \put(8.5,36.5){\textbf{\footnotesize Pour}}
    \put(7.3,13){\textbf{\footnotesize Bottle}}
    \put(75,-4){\textbf{(b)}}
    \put(25,-4){\textbf{(a)}}

    \end{overpic}

    \caption{\textbf{Visual Attention.} \textbf{(a)} Attention heatmaps on geometries with ($\bm{w}$) and without ($\bm{w/o}$) the AffCoT. \textbf{(b)} Feature maps on interaction images with ($\bm{w}$) and without ($\bm{w/o}$) the ObjCoT.}
    \label{fig:ablation}
    \vspace{-10pt}
\end{figure}

To further demonstrate the effectiveness of AffCoT and ObjCoT, we visualize attention maps on object geometries and interaction images when one of them is removed. As shown in Fig. \ref{fig:ablation} (a), relying solely on object geometric attributes without AffCoT results in a model that only focuses on affordance seen in the training set and fails to reason analogically for unseen affordance. The same issue arises with partial object point clouds. Additionally, as shown in Fig. \ref{fig:ablation} (b), we visualize features from the interaction images, as the algorithm is not limited by additional classification heads. When AffCoT guides the model to identify the approximate affordance region in the interaction image, ObjCoT inference further localizes the key interaction part of the object rather than the entire object, \eg. knife-cut perceives the blade of the knife and kettle-pour perceives the spout of the kettle in the interaction image.

\begin{table}
\centering
\footnotesize
\renewcommand{\arraystretch}{1.}
\renewcommand{\tabcolsep}{3.pt}
\caption{{\textbf{Ablation studies.} Performance when not modeling MHACoT with affordance-head CoT (AffCoT.) and object-head CoT (ObjCoT.), CMAFM and not introducing MLLM fine-tuning (FT.). \ding{55} means without.}}
\label{table:ablation}
\begin{tabular}{c|c|c|cccc}
\toprule
{} &\textbf{Metrics}   & \cellcolor{mygray}\textbf{Ours} & \textbf{\ding{55} AffCoT.} & \textbf{\ding{55} ObjCoT.} & \textbf{\ding{55} CMAFM} & \textbf{\ding{55} FT.} \\ \midrule

\multirow{4}{*}{\rotatebox{90}{\textbf{Seen}}} 
{}&\textbf{AUC}  & \cellcolor{mygray}91.99  & 90.88  & 90.13  & 89.52  & 88.75\\
{}&\textbf{aIOU} & \cellcolor{mygray}38.03  & 36.94  & 36.55  & 29.48  & 35.19\\
{}&\textbf{SIM}  & \cellcolor{mygray}0.676   & 0.659   & 0.651 & 0.590   & 0.625\\
{}&\textbf{MAE}  & \cellcolor{mygray}0.067  & 0.069  & 0.071  & 0.078  & 0.075\\
\cmidrule{1-7}
\multirow{4}{*}{\rotatebox{90}{\parbox{1cm}{\centering \textbf{Unseen}\\\textbf{Object}}}}
{}&\textbf{AUC}  & \cellcolor{mygray}79.57  & 74.58  & 75.87  & 78.42  & 77.83\\
{}&\textbf{aIOU} & \cellcolor{mygray}20.16  & 18.50  & 18.56  & 16.62  & 17.07\\
{}&\textbf{SIM}  & \cellcolor{mygray}0.402   & 0.390   & 0.383  & 0.349 & 0.374\\
{}&\textbf{MAE}  & \cellcolor{mygray}0.109  & 0.111  & 0.120  & 0.119  & 0.118\\
\cmidrule{1-7}
\multirow{4}{*}{\rotatebox{90}{\parbox{1.35cm}{\centering \textbf{Unseen}\\\textbf{Affordance}}}} 
{}&\textbf{AUC}  & \cellcolor{mygray}69.81  & 67.18  & 64.69  & 63.00  & 66.49\\
{}&\textbf{aIOU} & \cellcolor{mygray}12.05  & 10.93  & 8.81  & 6.24  & 10.06\\
{}&\textbf{SIM}  & \cellcolor{mygray}0.290  & 0.287   & 0.254   & 0.235 & 0.256\\
{}&\textbf{MAE}  & \cellcolor{mygray}0.127  & 0.164  & 0.133  & 0.128  & 0.134\\
\bottomrule
\end{tabular}
\vspace{-10pt}
\end{table}

\begin{figure}
    \centering
    \small
    \begin{overpic}[width=1.\linewidth]{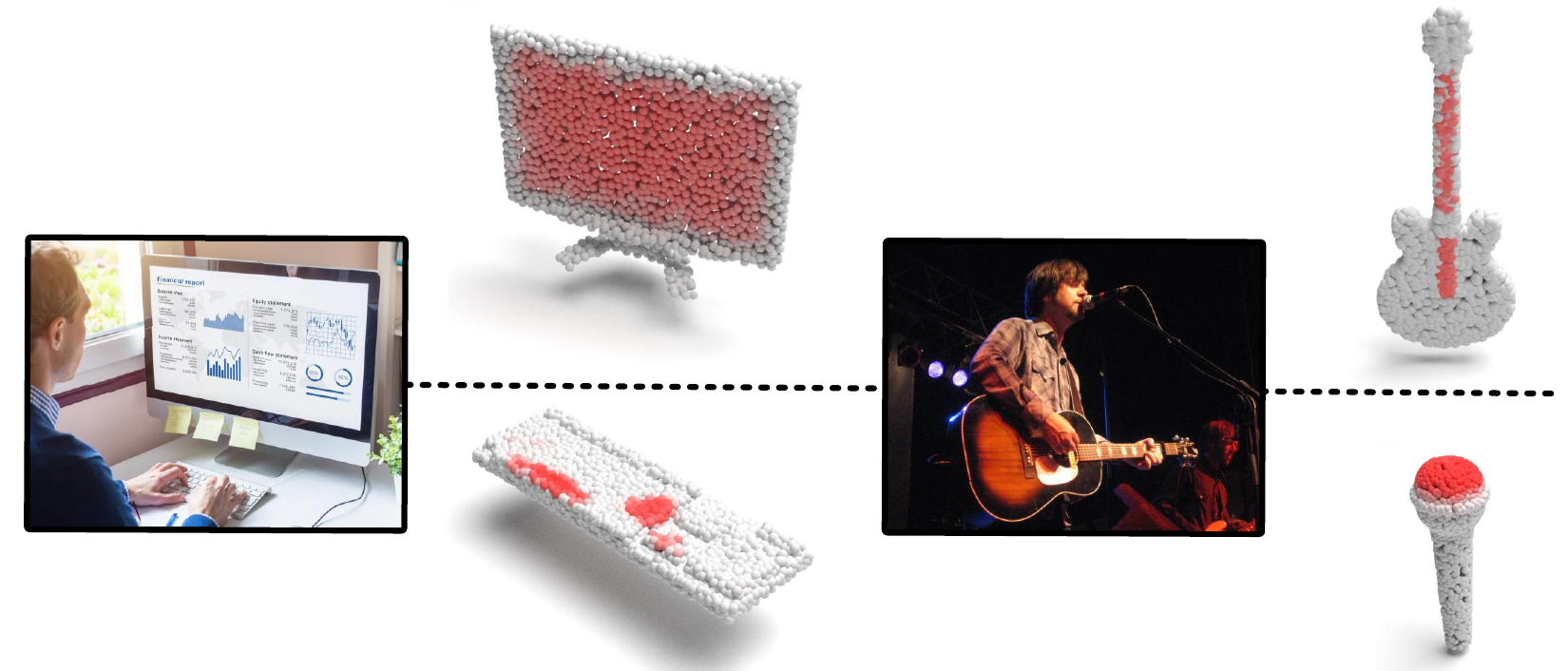}
    
    \put(9,32.5){\textbf{\footnotesize Screen}}
    \put(8.5,29){\textbf{\footnotesize Display}}
    \put(6.5,5.5){\textbf{\footnotesize Keyboard}}
    \put(9.5,2){\textbf{\footnotesize Press}}
    
    \put(63.5,32.5){\textbf{\footnotesize Guitar}}
    \put(65.5,29){\textbf{\footnotesize Play}}
    \put(60,5.5){\textbf{\footnotesize Microphone}}
    \put(64.8,2){\textbf{\footnotesize Speak}}
    
    \end{overpic}
    \vspace{-10pt}
    \caption{\textbf{Multiple Objects.} The anticipations for multiple objects with the same interaction image that contain different interactions.}
    \label{fig:multiobject}
  
\end{figure}
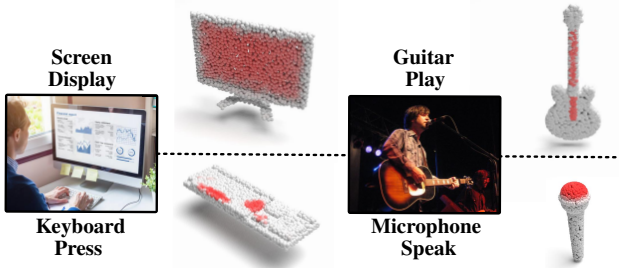

\subsection{Performance Analysis}
We conducted experiments separately for Seen, Unseen object, and Unseen affordance partitions, enabling a deeper analysis of the model's performance in various contexts.

\noindent\textbf{Multiple Objects.} In the case of humans interacting with distinct objects at the same time, the model needs to have the ability to understand interactions with different objects. As shown in Fig. \ref{fig:multiobject}, when reasoning about different objects of the same interaction image, the model can pinpoint the object affordance region.

\begin{figure}
    \centering
    \small
    \begin{overpic}[width=1.\linewidth]{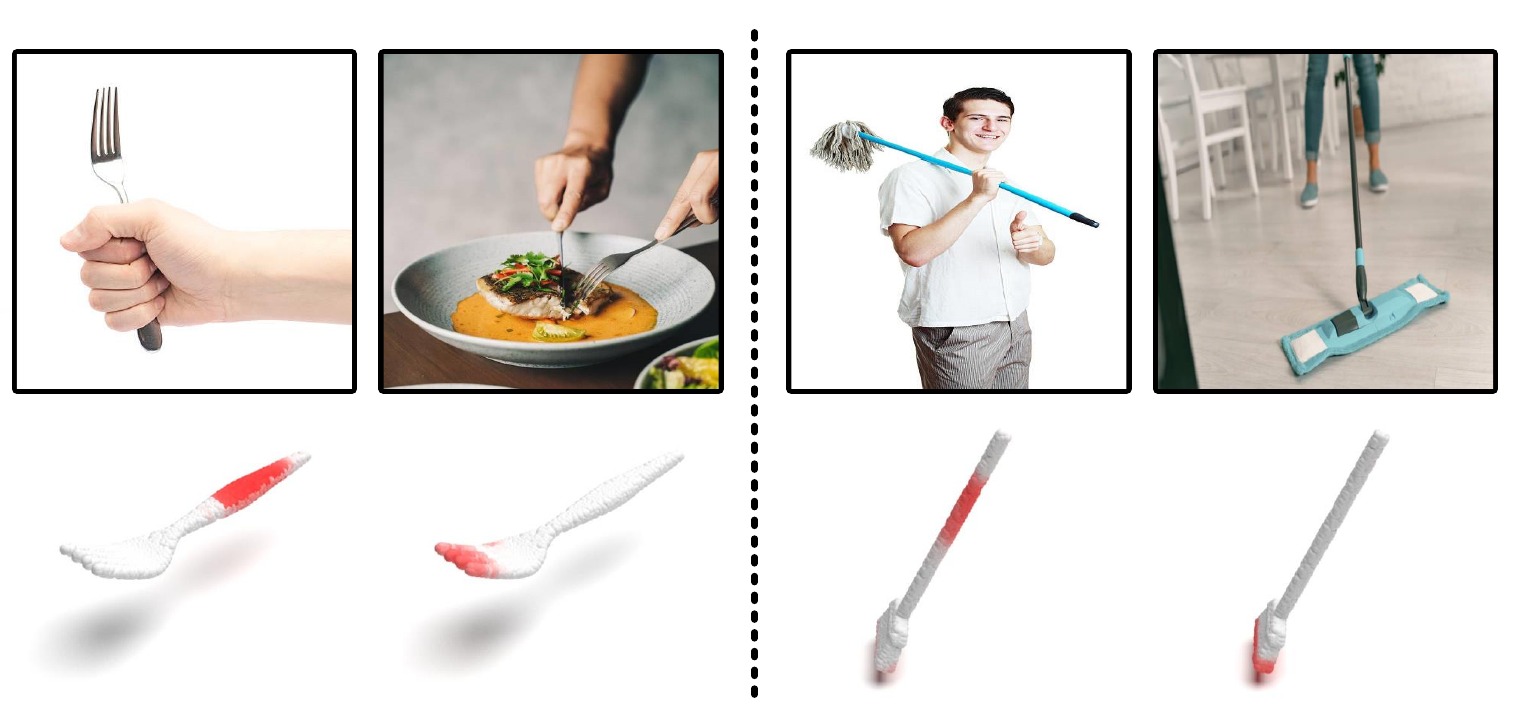}
    \put(20.5,45){\textbf{\footnotesize Fork}}
    \put(71.2,45){\textbf{\footnotesize Mop}}

    \put(3,-2.5){\textbf{\footnotesize Wrapgrasp}}
    \put(33,-2.5){\textbf{\footnotesize Stab}}
    \put(55,-2.5){\textbf{\footnotesize Wrapgrasp}}
    \put(83,-2.5){\textbf{\footnotesize Clean}}
    \end{overpic}
    \caption{\textbf{Multiple Affordances.} The anticipations for multiple affordances with the same point cloud for different interactions.}
    \vspace{-5pt}
    \label{fig:multiaff}
\end{figure}

\noindent\textbf{Multiple Affordances.} To verify whether the model reasons about the 3D object affordance regions based on the understanding of interaction images, we use the model to infer distinct affordances for the same objects, as shown in Fig. \ref{fig:multiaff}. The results demonstrate that, for the same object, the model outputs different results depending on the interaction, and the localized 3D object affordance regions are consistent with the interactions depicted in the 2D images.
%This indicates that the model reasons about the interaction details and localizes the affordance regions based on the interaction images, rather than relying solely on a direct category-driven mapping.

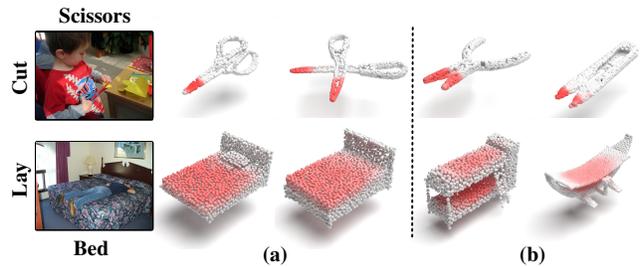
\begin{figure}
    \centering
    \small
    \vspace{5pt}
    \begin{overpic}[width=1.\linewidth]{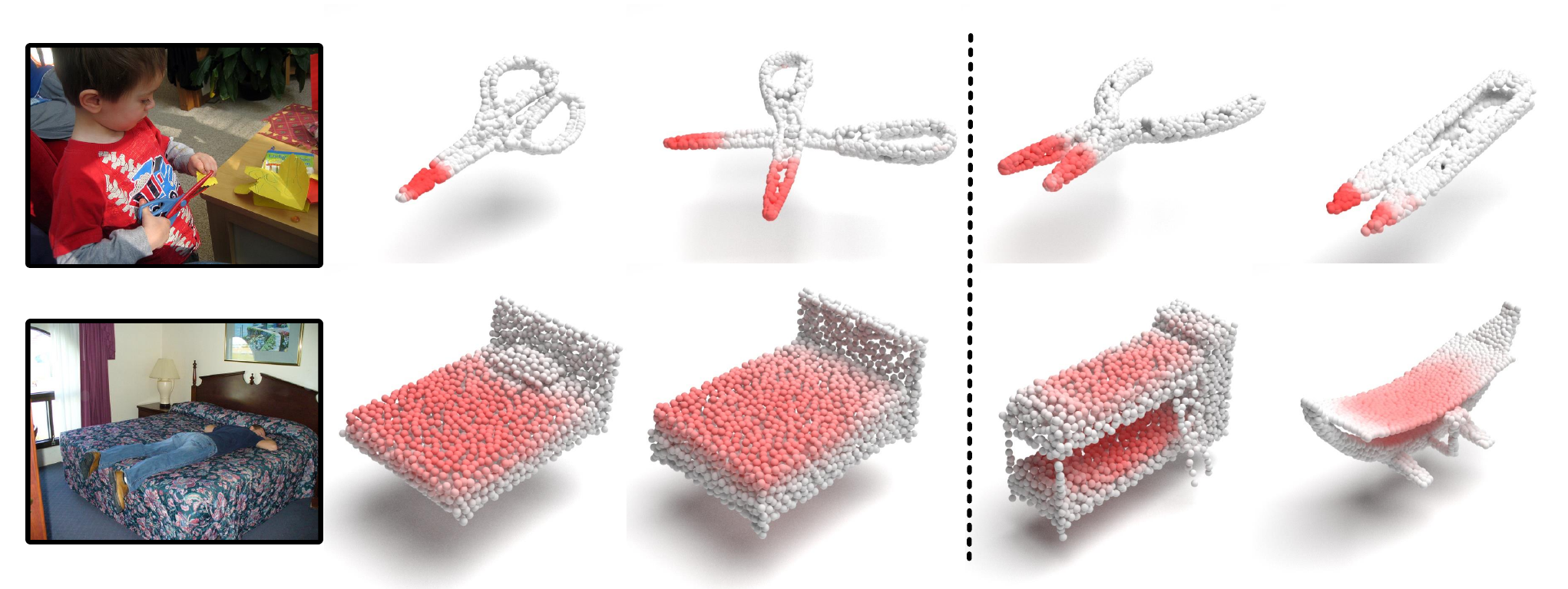}

    \put(-2,24.5){\rotatebox{90}{\textbf{\footnotesize Cut}}}
    \put(-2,8.2){\rotatebox{90}{\textbf{\footnotesize Lay}}}
    \put(5.5,36){\textbf{\footnotesize Scissors}}
    \put(7.8,-1){\textbf{\footnotesize Bed}}
    
    \put(38,-2){\textbf{\footnotesize (a)}}
    \put(79,-2){\textbf{\footnotesize (b)}}
    
    \end{overpic}
    \caption{\textbf{Multiple Instances.} \textbf{(a)} Similar geometric instances. \textbf{(b)} Different geometric instances.}
    
    \label{fig:multipoint}
\end{figure}

\noindent\textbf{Multiple Instances.} To assess the generalization and robustness of the model, we conduct an experiment to validate its understanding of different point cloud instances with the same object category, as shown in Fig. \ref{fig:multipoint}. The results demonstrate that the model can not only accurately locate point clouds that are highly similar to the shapes of the interacting objects in the images (Fig. \ref{fig:multipoint} (a)), but also can effectively locate the same affordance regions in point clouds of the same category, even with significant geometric variations (Fig. \ref{fig:multipoint} (b)), demonstrating its ability to generalize affordance grounding and maintain robustness under geometric variations.

\section{Conclusion}
We present grounding 3D object affordance in an open-vocabulary fashion, which reasons from interaction images, extrapolating from predefined sample space and generalize to unseen scenarios. To achieve so, we propose a novel framework to  utilize multi-head affordance chain-of-thought reasoning, excavating invariant geometric properties and analogous reasoning about potential interactions, with the alignment of cross-modal features to localize 3D object affordance region. Furthermore, We introduce the largest 3D object affordance dataset PIADv2, which contains 15\textit{K} interaction images and over 38\textit{K} 3D objects with annotations. Extensive experiments demonstrate the effectiveness and superiority of GREAT. It supports affordance understanding in open scenarios, potentially enhancing robots' autonomous interaction in unknown environments. We believe it could offer fresh insights and promote research in the area of visual affordance understanding.

\noindent\textbf{Limitations and Future Work.}
The limitation of GREAT lies in the high computational complexity of its multi-step reasoning, which can become a bottleneck in large-scale or real-time applications. In future work, we aim to create inference-specific datasets and use them to distill multi-modal models into specialized knowledge domains, enabling faster and more efficient performance in real-world.

\noindent\textbf{Acknowledgments}. This work is supported by the National Natural Science Foundation of China (NSFC) under Grants 62306295, 62225207 and 62436008.

{
    \small
    \bibliographystyle{ieeenat_fullname}
    \bibliography{main}
}

\end{document}